\documentclass{article} 
\usepackage{iclr2016_conference,times}
\usepackage{url}
\usepackage{graphicx}
\usepackage{subfig}
\usepackage{amsmath}
\usepackage{floatrow}
\usepackage{amsfonts}
\usepackage{multirow}
\usepackage{float}
\usepackage{algorithm}
\usepackage{algpseudocode}
\usepackage[innercaption]{sidecap}
\usepackage{pifont}
\sidecaptionvpos{figure}{t}
\iclrfinalcopy

\title{\centering Neural Programmer: Inducing Latent Programs with Gradient Descent}

\author{
Arvind Neelakantan\thanks{Work done during an internship at Google.} \\
University of Massachusetts Amherst\\
\texttt{arvind@cs.umass.edu} \\
\And
Quoc V. Le \\
{Google Brain} \\
\texttt{qvl@google.com} \\
\And
Ilya Sutskever \\
{Google Brain} \\
\texttt{ilyasu@google.com} \\
}

\begin{document}
\maketitle
\begin{abstract}
Deep neural networks have achieved impressive supervised
classification performance in many tasks including image recognition, speech
recognition, and sequence to sequence learning. However, this success has not been translated to applications like question answering
that may involve complex arithmetic and logic reasoning. A major
limitation of these models is in their inability to learn even simple
arithmetic and logic operations. For example, it has been shown that
neural networks fail to learn to add two binary numbers reliably. In
this work, we propose {\it Neural Programmer}, a neural network augmented with a small set of basic
arithmetic and logic operations that can be trained end-to-end using backpropagation. Neural Programmer can call these
augmented operations over several steps, thereby inducing
compositional programs that are more complex than the built-in
operations.
The model learns from a weak supervision signal which is the result of
execution of the correct program, hence it does not require expensive
annotation of the correct program itself. The decisions of what
operations to call, and what data segments to apply to are inferred by
Neural Programmer. Such decisions, during training, are done in a
differentiable fashion so that the entire network can be trained
jointly by gradient descent. We find that training the model is
difficult, but it can be greatly improved by adding random noise to
the gradient. On a fairly complex synthetic table-comprehension
dataset, traditional recurrent networks and attentional models perform
poorly while Neural Programmer typically obtains nearly perfect
accuracy.
\end{abstract}

\section{Introduction}
The past few years have seen the tremendous success of deep neural
networks (DNNs) in a variety of supervised classification tasks
starting with image recognition \citep{KrizhevskySH12} and speech
recognition \citep{38131} where the DNNs act on a fixed-length input
and output. More recently, this success has been translated into
applications that involve a variable-length sequence as input and/or
output such as machine translation
\citep{SutskeverVL14,BahdanauCB14,luong2014addressing}, image
captioning \citep{VinyalsTBE15,XuBKCCSZB15}, conversational
modeling~\citep{shang2015neural,vinyals2015neural}, end-to-end
Q\&A~\citep{sukhbaatar2015weakly,DBLP:journals/corr/PengLLW15,DBLP:journals/corr/HermannKGEKSB15},
and end-to-end speech recognition
\citep{graves2014towards,HannunCCCDEPSSCN14,ChanJLV15,BahdanauCSBB15}.

While these results strongly indicate that DNN models are capable of
learning the fuzzy underlying patterns in the data, they have not had
similar impact in applications that involve crisp reasoning. A major
limitation of these models is in their inability to learn even simple
arithmetic and logic operations. For example, \citet{JoulinM15} show
that recurrent neural networks (RNNs) fail at the task of adding two
binary numbers even when the result has less than 10 bits. This makes
existing DNN models unsuitable for downstream applications that
require complex reasoning, e.g., natural language question
answering. For example, to answer the question ``how many states
border Texas?'' (see \citet{Zettlemoyer05learningto}), the algorithm
has to perform an act of counting in a table which is something that a
neural network is not yet good at.

A fairly common method for solving these problems is {\it program
  induction} where the goal is to find a program (in SQL or some
high-level languages) that can correctly solve the task. An
application of these models is in {\it semantic parsing} where the
task is to build a natural language interface to a structured database
\citep{zelle:aaai96}. This problem is often formulated as mapping a
natural language question to an executable query.


A drawback of existing methods in semantic parsing is that they are
difficult to train and require a great deal of human supervision. As
the space over programs is non-smooth, it is difficult to apply simple
gradient descent; most often, gradient descent is augmented with a
complex search procedure, such as
sampling~\citep{liang2010learning}. To further simplify training, the
algorithmic designers have to manually add more supervision signals to
the models in the form of annotation of the complete program for every
question \citep{Zettlemoyer05learningto} or a domain-specific grammar
\citep{Liang:2011}. For example, designing grammars that contain rules
to associate lexical items to the correct operations, e.g., the word
``largest'' to the operation ``argmax'', or to produce syntactically
valid programs, e.g., disallow the program $>= dog$. The role of
hand-crafted grammars is crucial in semantic parsing yet also limits
its general applicability to many different domains.  In a recent work
by \citet{WangBL15} to build semantic parsers for $7$ domains, the
authors hand engineer a separate grammar for each domain.

\begin{figure}[h!]
  \includegraphics[scale=.25]{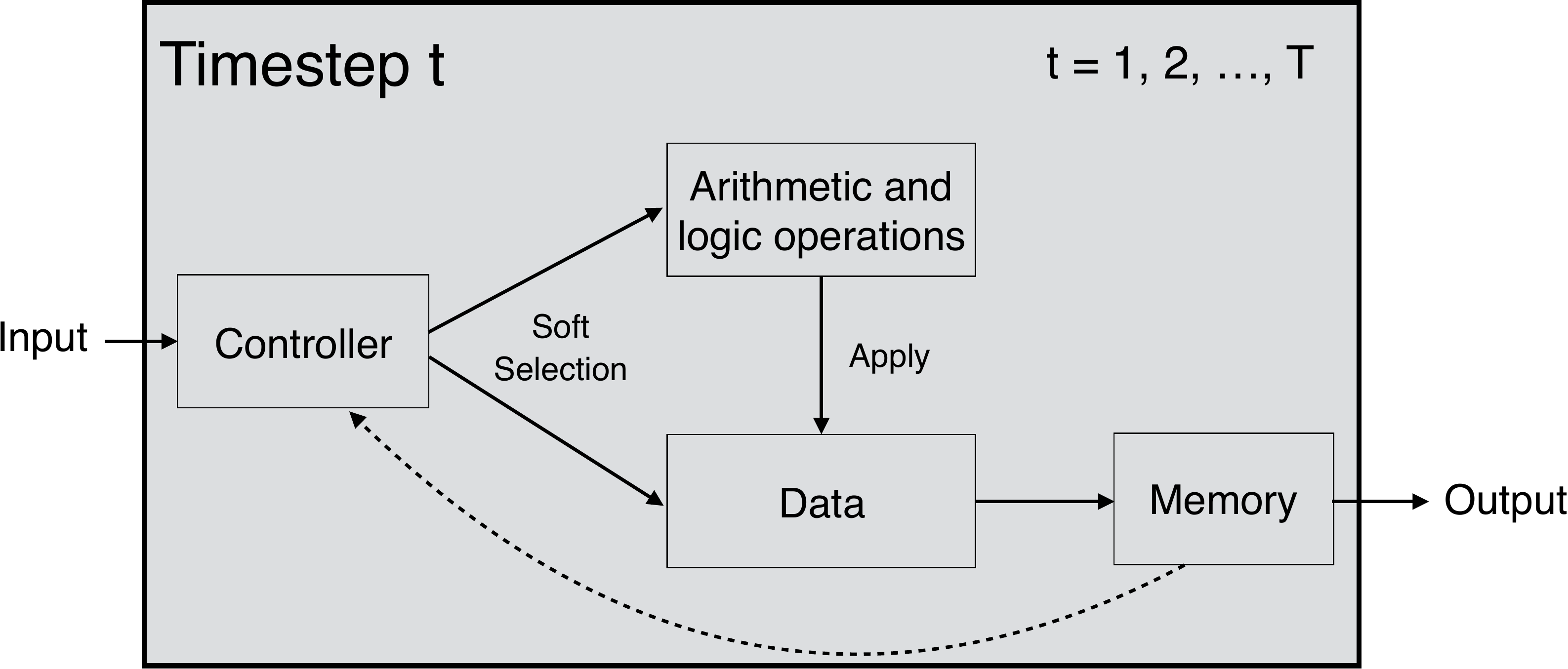}
  \caption{The architecture of Neural Programmer, a neural network
    augmented with arithmetic and logic operations. The controller
    selects the operation and the data segment. The memory stores the
    output of the operations applied to the data segments and the
    previous actions taken by the controller. The controller runs for
    several steps thereby inducing compositional programs that are
    more complex than the built-in operations. The dotted line
    indicates that the controller uses information in the memory to
    make decisions in the next time step.} \label{intro}
\end{figure}

The goal of this work is to develop a model that does not require
substantial human supervision and is broadly applicable across
different domains, data sources and natural languages. We propose {\it
  Neural Programmer} (Figure \ref{intro}), a neural network augmented with a small set of basic
arithmetic and logic operations that can be trained end-to-end using backpropagation.  In our formulation, the neural
network can run several steps using a recurrent neural network. At
each step, it can select a segment in the data source and a particular
operation to apply to that segment. The neural network propagates
these outputs forward at every step to form the final, more
complicated output. Using the target output, we can adjust the network
to select the right data segments and operations, thereby inducing the
correct program. Key to our approach is that the selection process
(for the data source and operations) is done in a differentiable
fashion (i.e., soft selection or attention), so that the whole neural
network can be trained jointly by gradient descent.  At test time, we
replace soft selection with hard selection.

By combining neural network with mathematical operations, we can
utilize both the fuzzy pattern matching capabilities of deep networks
and the crisp algorithmic power of traditional programmable
computers. This approach of using an augmented logic and arithmetic
component is reminiscent of the idea of using an ALU (arithmetic and
logic unit) in a conventional computer~\citep{von1993first}. It is
loosely related to the symbolic numerical processing abilities
exhibited in the intraparietal sulcus (IPS) area of the
brain~\citep{piazza2004tuning,Cantlon06,kucian2006impaired,fias2007processing,dastjerdi2013numerical}. Our
work is also inspired by the success of the soft attention mechanism
\citep{BahdanauCB14} and its application in learning a neural network
to control an additional memory component
\citep{GravesWD14,sukhbaatar2015weakly}.

Neural Programmer has two attractive properties. First, it learns from
a weak supervision signal which is the result of execution of the
correct program. It does not require the expensive annotation of the
correct program for the training examples.  The human supervision
effort is in the form of question, data source and answer
triples. Second, Neural Programmer does not require additional rules
to guide the program search, making it a general framework.  With
Neural Programmer, the algorithmic designer only defines a list of
basic operations which requires lesser human effort than in previous
program induction techniques.


We experiment with a synthetic table-comprehension dataset, consisting
of questions with a wide range of difficulty levels. Examples of
natural language translated queries include ``print elements in column H whose field in column C is greater than 50 and
field in column E is less than 20?''  or ``what is the difference
between sum of elements in column A and number of rows in the
table?''. We find that LSTM recurrent networks~\citep{Hochreiter:1997}
and LSTM models with attention \citep{BahdanauCB14} do not work
well. Neural Programmer, however, can completely solve this task or
achieve greater than 99\% accuracy on most cases by inducing the
required latent program. We find that training the model is difficult,
but it can be greatly improved by injecting random Gaussian noise to
the gradient~\citep{WellingT11,noise} which enhances the generalization
ability of the Neural Programmer.

\section{Neural Programmer}
\begin{figure}[h!]
  \caption{ An implementation of Neural Programmer for the task of
    question answering on tables. The output of the model at time step
    $t$ is obtained by applying the operations on the data segments
    weighted by their probabilities. The final output of the model is
    the output at time step $T$. The dotted line indicates the input
    to the history RNN at step t+1.  } \label{main:high-level}
  \includegraphics[scale=.25]{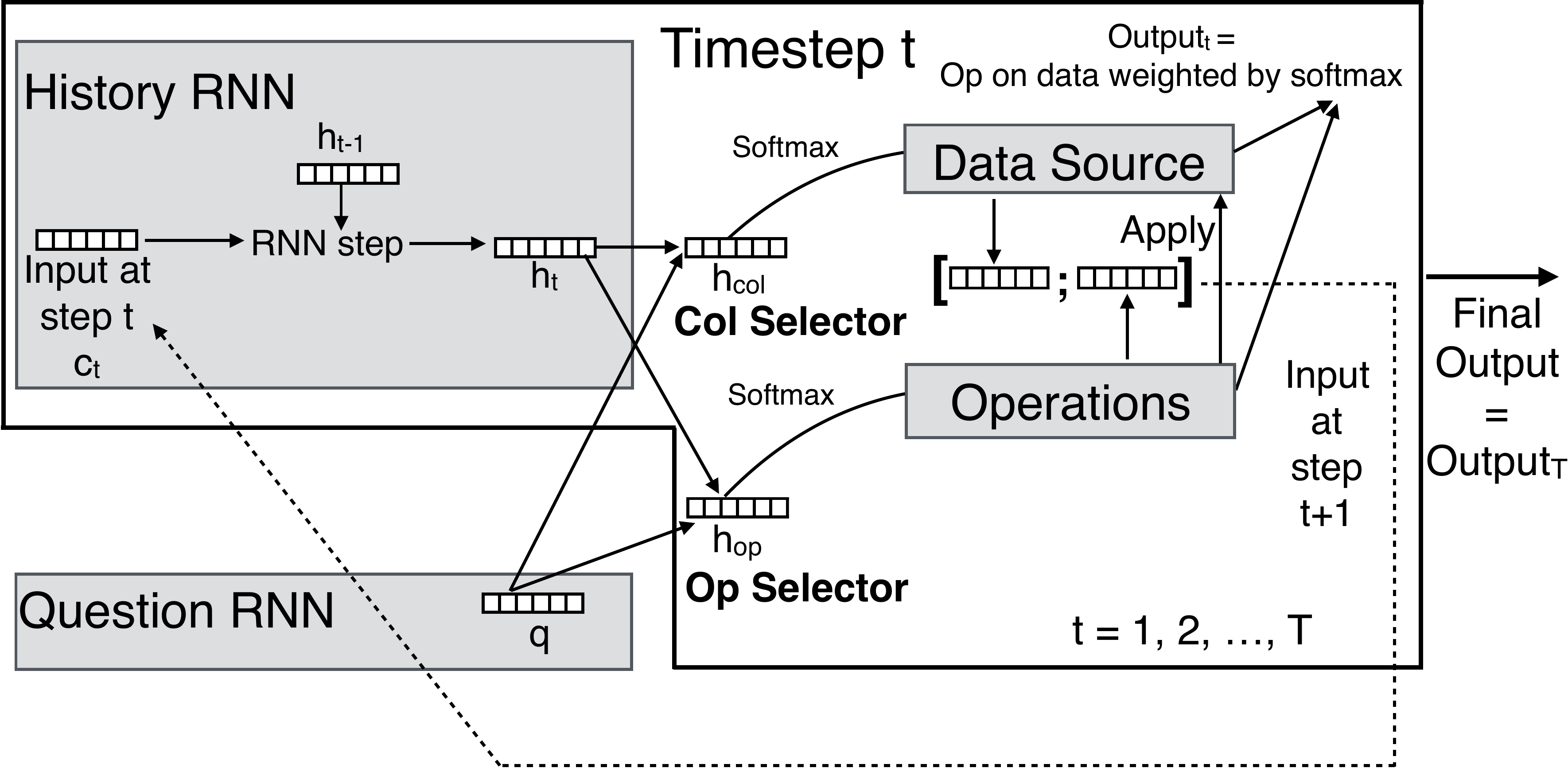}
\end{figure}

Even though our model is quite general, in this paper, we apply
Neural Programmer to the task of question answering on tables, a task
that has not been previously attempted by neural networks. In our
implementation for this task, Neural Programmer is run for a total of
$T$ time steps chosen in advance to induce compositional programs of
up to $T$ operations. The model consists of four modules:
\begin{itemize} 
\item A question Recurrent Neural Network (RNN) to process the input
  question,
\item A selector to assign two probability distributions at every step, one over the set of
operations and the other over the data segments, 
\item A list of operations
that the model can apply and, 
\item A history RNN to remember the previous
operations and data segments selected by the model till the current time step.
\end{itemize}
These four modules are also shown in Figure~\ref{main:high-level}. The
history RNN combined with the selector module functions as the
controller in this case.  Information about each component is
discussed in the next sections.

Apart from the list of operations, all the other modules are learned
using gradient descent on a training set consisting of triples, where
each triple has a question, a data source and an answer. We assume
that the data source is in the form of a table, $\mathit{table} \in
\mathbb{R}^{M \times C}$, containing $M$ rows and $C$ columns ($M$ and
$C$ can vary amongst examples). The data segments in our
experiments are the columns, where each column also has a column name.

\subsection{Question Module}
\label{question}
\begin{figure}[h!]
  \includegraphics[scale=.30]{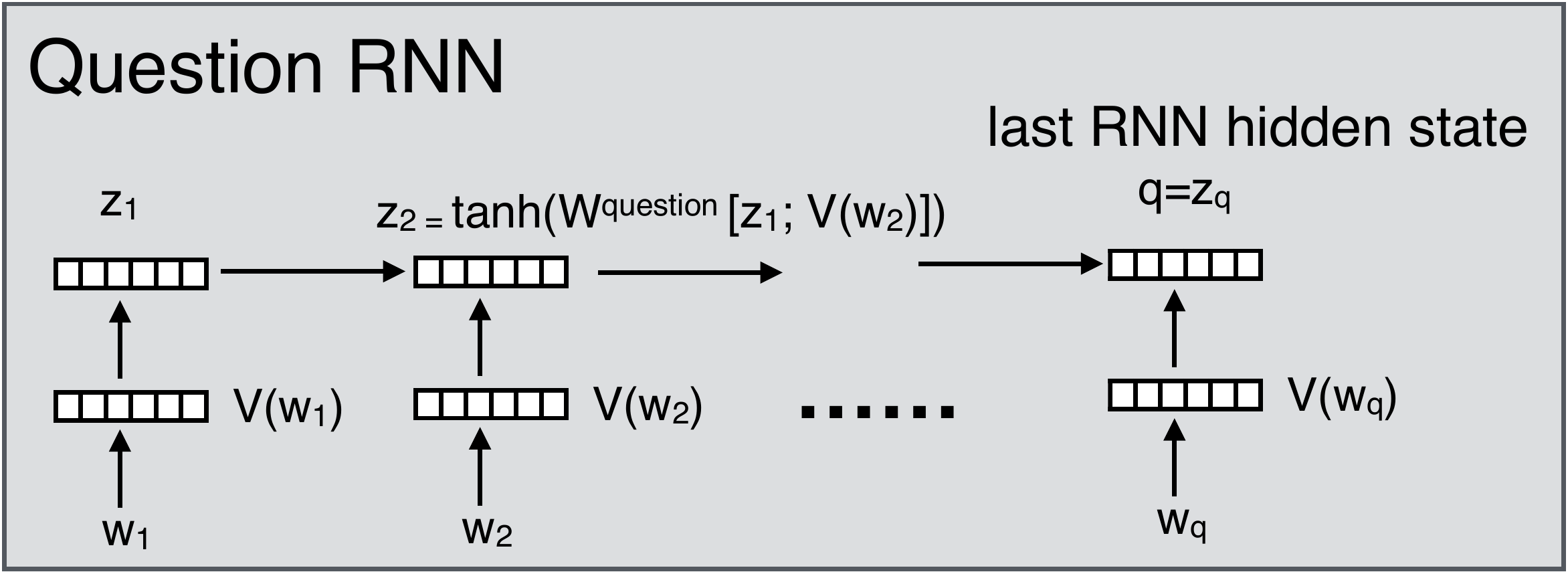}
  \caption{The question module to process the input question. $q=z_q$
    denotes the question representation used by Neural Programmer.
  } \label{main:question}
\end{figure}

The question module converts the question tokens to a distributed
representation. In the basic version of our model, we use a simple RNN
\citep{werbos:bptt} parameterized by $W^{question}$ and the last
hidden state of the RNN is used as the question representation (Figure
\ref{main:question}).

Consider an input question containing $Q$ words $\{w_{1}, w_{2},
\ldots, w_{Q}\}$, the question module performs the following
computations:
\begin{align*}
z_{i} = \tanh(W^{\mathit{question}}[z_{i-1}; V(w_{i})]), \forall i={1, 2, \ldots, Q}
\end{align*}
where $V(w_{i}) \in \mathbb{R}^{d}$ represents the embedded
representation of the word $w_{i}$, $[a;b] \in \mathbb{R}^{2d}$
represents the concatenation of two vectors $a, b \in \mathbb{R}^{d}$,
$W^{\mathit{question}} \in \mathbb{R}^{d \times 2d}$ is the recurrent matrix of
the question RNN, 
$\tanh$ is the element-wise non-linearity
function and $z_Q \in \mathbb{R}^{d}$ is the representation of the
question. We set $z_0$  to $[0]^{d}$. We pre-process the question by
removing numbers from it and storing the numbers in a separate
list. Along with the numbers we store the word that appeared to the
left of it in the question which is useful to compute the pivot
values for the comparison operations described in Section \ref{ops}.

For tasks that involve longer questions, we use a bidirectional RNN
since we find that a simple unidirectional RNN has trouble remembering
the beginning of the question. When the bidirectional RNN is used, the
question representation is obtained by concatenating the last hidden
states of the two-ends of the bidirectional RNNs. The question
representation is denoted by $q$.

\subsection{Selector}
\label{selector}
The selector produces two probability distributions at every time step
$t \; (t = 1, 2, \ldots, T)$: one probablity distribution over the set
of operations and another probability distribution over the set of
columns. The inputs to the selector are the question representation
($q \in \mathbb{R}^{d}$) from the question module and the output of
the history RNN (described in Section \ref{history}) at time step $t$
($h_{t} \in \mathbb{R}^{d}$) which stores information about the
operations and columns selected by the model up to the previous step.

Each operation is represented using a $d$-dimensional vector. Let the
number of operations be $O$ and let $U \in \mathbb{R}^{O \times D}$ be
the matrix storing the representations of the operations.

\begin{figure}[h!]
  \includegraphics[scale=.26]{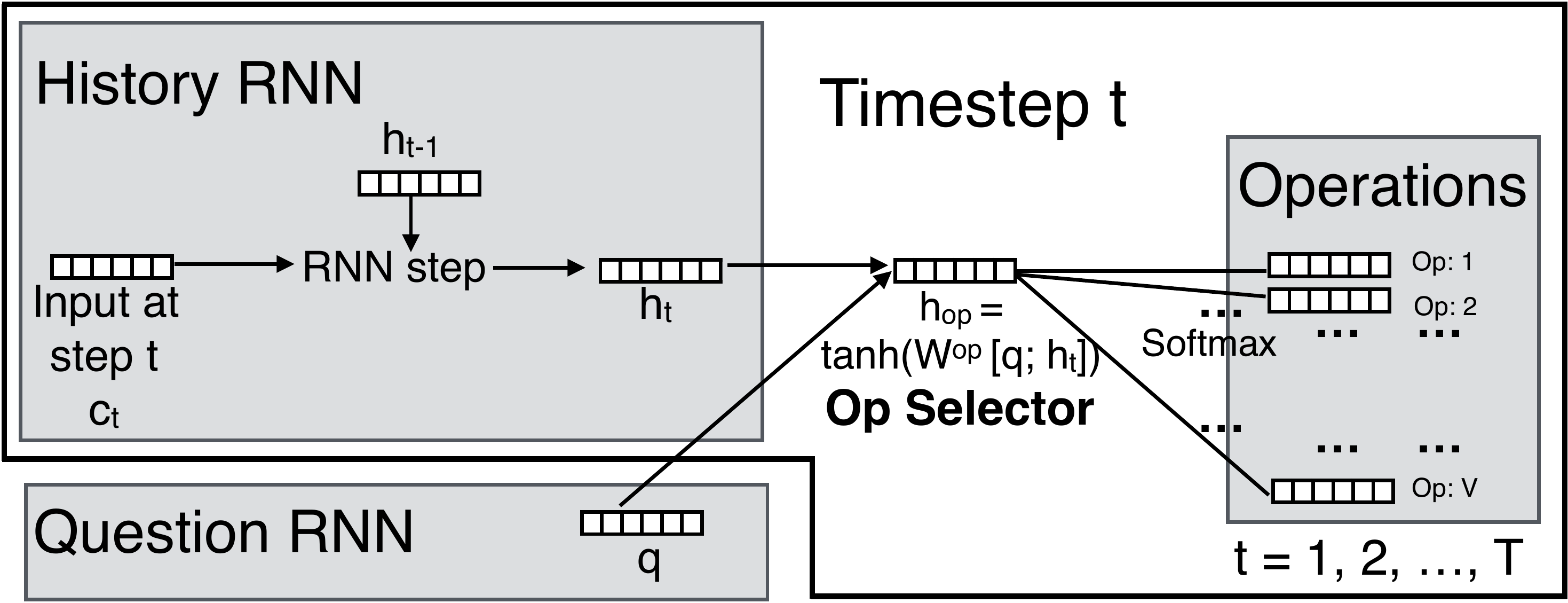}
  \caption{
     Operation selection at time step $t$ where the selector assigns a probability distribution over the set of operations.
    } \label{main:op}
\end{figure}

\noindent {\bf Operation Selection} is performed by:
\begin{align*}
 \alpha^{op}_{t} = \mathit{softmax}(U \tanh(W^{op}[q;h_t]))
\end{align*}
where $W^{op} \in \mathbb{R}^{d \times 2d}$ is the parameter matrix of
the operation selector that produces the probability distribution
$\alpha^{op}_{t} \in [0,1]^{O}$ over the set of operations (Figure
\ref{main:op}).

The selector also produces a probability distribution over the columns
at every time step. We obtain vector representations for the column
names using the parameters in the question module (Section
\ref{question}) by word embedding or an RNN phrase embedding. Let $P
\in \mathbb{R}^{C \times D}$ be the matrix storing the representations
of the column names.  

\begin{figure}[h!]
  \includegraphics[scale=.26]{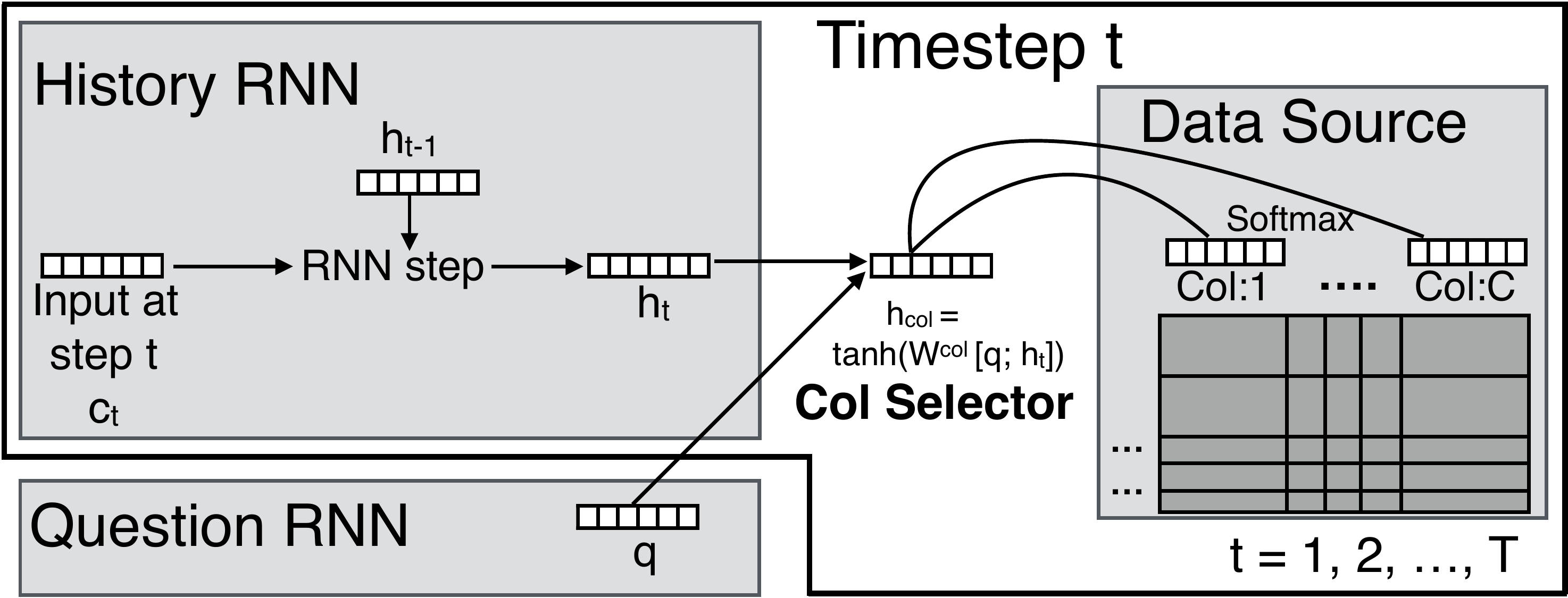}
  \caption{
     Data selection at time step $t$ where the selector assigns a probability distribution over the set of columns.
    } \label{main:data}
\end{figure}

{\bf Data Selection} is performed by:

\begin{align*}
 \alpha^{col}_{t} = \mathit{softmax}(P \tanh(W^{col}[q;h_t]))
\end{align*}
where $W^{col} \in \mathbb{R}^{d \times 2d}$ is the parameter matrix
of the column selector that produces the probability distribution
$\alpha^{col}_{t} \in [0,1]^{C}$ over the set of columns (Figure
\ref{main:data}).

\subsection{Operations}
\label{ops}
Neural Programmer currently supports two types of outputs: a) a scalar
output, and b) a list of items selected from the table (i.e., table
lookup).\footnote{It is trivial to extend the model to support general
  text responses by adding a decoder RNN to generate text sentences.}
The first type of output is for questions of type ``Sum of elements in
column C'' while the second type of output is for questions of type
``Print elements in column A that are greater than 50.'' To facilitate
this, the model maintains two kinds of output variables at every step
$t$, $\mathit{scalar\_answer_{t}} \in \mathbb{R}$ and $\mathit{lookup\_answer_{t}} \in
[0,1]^{M \times C}$. The output $\mathit{lookup\_answer_{t} (i, j)}$ stores the
probability that the element $(i,j)$ in the table is part of the
output. The final output of the model is $\mathit{scalar\_answer_{T}}$ or
$\mathit{lookup\_answer_{T}}$ depending on whichever of the two is updated after
$T$ time steps.  Apart from the two output variables, the model
maintains an additional  variable $\mathit{row\_select_t} \in [0,1]^{M}$ that is
updated at every time step. The variables $\mathit{row\_select_t[i]} (\forall
i=1,2,\ldots,M$) maintain the probability of selecting row $i$ and
allows the model to dynamically select a subset of rows within a column. The output is initialized to zero while the $\mathit{row\_select}$
variable is initialized to $[1]^{M}$.

Key to Neural Programmer is the built-in operations, which have access
to the outputs of the model at every time step before the current time
step $t$, i.e., the operations have access to
$(\mathit{scalar\_answer}_{i},\mathit{lookup\_answer}_{i}), \forall
i={1,2,\ldots,t-1}$. This enables the model to build powerful
compositional programs.

It is important to design the operations such that they can work with probabilistic
row and column selection so that the model is differentiable. Table
\ref{operations} shows the list of operations built into the model
along with their definitions. The reset operation can be selected any
number of times which when required allows the model to induce
programs whose complexity is less than $T$ steps.

\begin{center}
    \begin{table}[h!]
     \small
    \begin{tabular}{| l | l | l |}
    \hline
    Type & Operation & Definition \\ \hline \hline
    {\multirow{2}{*} {Aggregate}} & Sum & $\mathit{sum_t}[j] = \sum\limits_{i=1}^{M} \mathit{row\_select_{t-1}}[i] * \mathit{table}[i][j], \forall j = 1, 2, \ldots, C$     \\ 
                               & Count & $\mathit{count_t} = \sum\limits_{i=1}^{M} \mathit{row\_select_{t-1}}[i] $ \\ \hline
    Arithmetic & Difference & $\mathit{diff_t} = \mathit{scalar\_output}_{t - 3} - \mathit{scalar\_output}_{t - 1}$ \\ \hline
    {\multirow{2}{*} {Comparison}} & Greater & $g_t[i][j] = \mathit{table}[i][j] > \mathit{pivot_{g}}, \forall (i, j), i = 1, \ldots, M, j = 1, \ldots, C$    \\ 
                               & Lesser & $l_t[i][j] = \mathit{table[i][j]} < \mathit{pivot_{l}}, \forall (i, j), i = 1, \ldots, M, j = 1, \ldots, C$  \\ \hline
    {\multirow{2}{*} {Logic}} & And & $\mathit{and}_t[i] = min(\mathit{row\_select_{t-1}[i]}, \mathit{row\_select_{t-2}[i]}), \forall i = 1, 2, \ldots, M$   \\ 
                               & Or & $\mathit{or}_t[i] = max(\mathit{row\_select_{t-1}[i]}, \mathit{row\_select_{t-2}[i]}), \forall i = 1, 2, \ldots, M$   \\ \hline
    Assign Lookup & assign & $assign_t[i][j] = \mathit{row\_select_{t-1}[i]}, \forall (i, j) i=1,2,\ldots, M, j = 1, 2, \ldots, C$   \\ \hline
    Reset & Reset &  $\mathit{reset_t[i]} = 1, \forall i = 1, 2, \ldots, M$    \\ \hline
    \end{tabular}
    \caption{List of operations along with their definitions at time
      step $t$, $table \in \mathbb{R}^{M \times C}$ is the data source
      in the form of a table and $row\_select_t \in [0,1]^{M}$ functions as a
      row selector.}
    \label{operations}
    \end{table}
\end{center}

While the definitions of the operations are fairly straightforward,
comparison operations greater and lesser require a pivot value as
input (refer Table \ref{operations}), which appears in the question.  Let $qn_1, qn_2, \ldots, qn_N$
be the numbers that appear in the question.

For every comparison operation (greater and lesser), we compute its
pivot value by adding up all the numbers in the question each of them
weighted with the probabilities assigned to it computed using the
hidden vector at position to the left of the number,\footnote{This
  choice is made to reflect the common case in English where the pivot
  number is usually mentioned after the operation but it is trivial to
  extend to use hidden vectors both in the left and the right of the
  number.} and the operation's embedding vector. More precisely:
\begin{align*}
\beta_{\mathit{op}} = \mathit{softmax}(Z U(\mathit{op})) \\
\mathit{pivot_{op}} = \sum \limits_{i=1}^{N} \beta_{\mathit{op}}(i) qn_i 
\end{align*}
where $ U(\mathit{op}) \in \mathbb{R}^{d}$ is the vector representation of
operation $\mathit{op}$ ($\mathit{op} \in \{\text{greater}, \text{lesser}\}$) and $ Z \in \mathbb{R}^{N \times d}$ is the matrix
storing the hidden vectors of the question RNN at positions to the
left of the occurrence of the numbers.

By overloading the definition of $\alpha^{op}_t$ and $\alpha^{col}_t$,
let $\alpha^{op}_t(x)$ and $\alpha^{col}_t(j)$ denote the probability
assigned by the selector to operation $x$ ($x \in$ \{sum, count,
difference, greater, lesser, and, or, assign, reset\}) and column $j$
($\forall j = 1,2,\ldots,C$) at time step $t$ respectively.  

\begin{figure}[h!]
  \includegraphics[scale=.28]{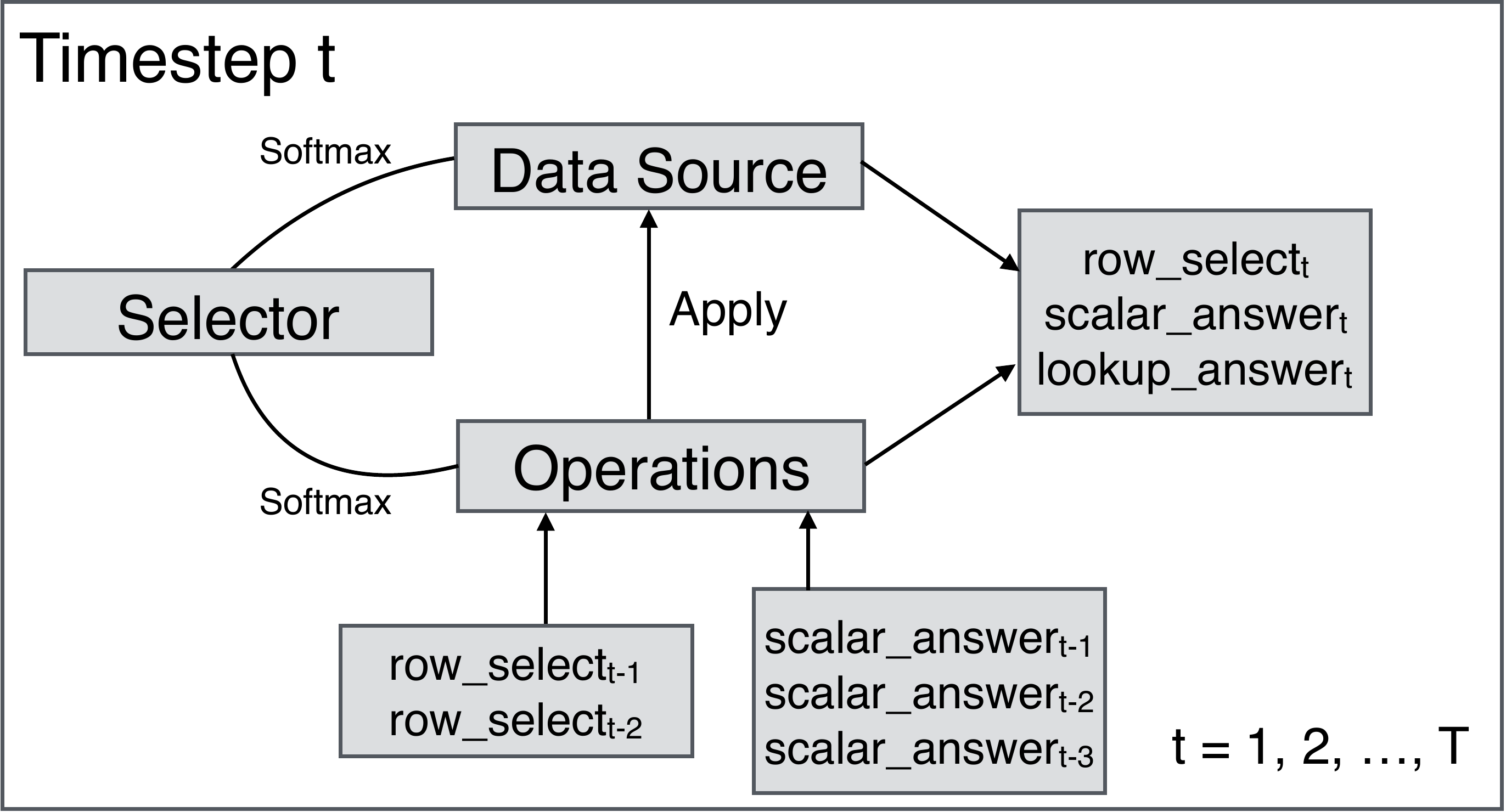}
  \caption{The output and row selector variables are obtained by applying the operations on the data segments and additively combining their outputs weighted using the probabilities assigned by the selector.} 
  \label{main:output}
\end{figure}

Figure \ref{main:output} show how the output and row selector
variables are computed. The output and row selector variables at a
step is obtained by additively combining the output of the individual
operations on the different data segments weighted with their
corresponding probabilities assigned by the model.

More formally, the output variables are given by:

\begin{align*}
  &\mathit{scalar\_answer_t} = \alpha^{op}_t(\text{count}) \mathit{count_t}  + \alpha^{op}_t(\text{difference}) \mathit{diff_t} + \sum \limits_{j=1}^C \alpha^{col}_t(j) \alpha^{op}_t(\text{sum}) \mathit{sum_t[j]}, \\
  &\mathit{lookup\_answer_t}[i][j] = \alpha^{col}_t(j) \alpha^{op}_t(\text{assign}) \mathit{assign_t}[i][j], \forall (i, j) i=1,2,\ldots, M, j = 1, 2, \ldots, C
\end{align*}
The row selector variable is given by:
\begin{align*}
 \mathit{row\_select_t[i]} = \alpha^{op}_t(\text{and}) & \mathit{and_t}[i]  + \alpha^{op}_t(\text{or}) \mathit{or_t}[i] + \alpha^{op}_t(\text{reset}) \mathit{reset_t}[i] + \\ &\sum \limits_{j=1}^C \alpha^{col}_t(j) (\alpha^{op}_t(\text{greater}) g_t[i][j] + \alpha^{op}_t(\text{lesser}) l_t[i][j]), \forall i = 1, \ldots, M
\end{align*}

It is important to note that other operations like {\it equal to, max,
  min, not} etc. can be built into this model easily. 
  
\subsubsection{Handling Text Entries}
\label{text}
So far, our disscusion has been only concerned with tables that have
numeric entries. In this section we describe how Neural Programmer
handles text entries in the input table. We assume a column can
contain either numeric or text entries. An example query is ``what is
the sum of elements in column B whose field in column C is word:1 and
field in column A is word:7?''. In other words, the query is looking
for text entries in the column that match specified words in the
questions. To answer these queries, we add a text match operation that
updates the row selector variable appropriately. In our
implementation, the parameters for vector representations of the
column's text entries are shared with the question module.

The text match operation uses a two-stage soft attention mechanism,
back and forth from the text entries to question module. In the
following, we explain its implementation in detail.

Let $TC_1, TC_2, \ldots, TC_K$ be the set of columns that each have
$M$ text entries and $A \in {M \times K \times d}$ store the vector
representations of the text entries. In the first stage, the question
representation coarsely selects the appropriate text entries through
the sigmoid operation.  Concretely, coarse selection, $B$, is given by
the sigmoid of dot product between vector representations for text
entries, $A$, and question representation, $q$:

\begin{align*}
 B[m][k] = \mathit{sigmoid} \left(\sum \limits_{p=1}^d A[m][k][p] \cdot q[p] \right) \; \; \forall (m,k) \; \; m = 1, \ldots, M, k = 1, \ldots, K  
\end{align*}

To obtain question-specific column representations, $D$, we use $B$ as
weighting factors to compute the weighted average of the vector
representations of the text entries in that column:

\begin{align*}
 D[k][p] = \frac{1}{M}\sum \limits_{m=1}^M \left( B[m][k] \cdot A[m][k][p] \right) \; \; \forall (k,p) \; \; k = 1, \ldots, K,  p = 1, \ldots, d
\end{align*}

To allow different words in the question to be matched to the corresponding columns (e.g., match word:1 in column C and match word:7 in column A for question ``what is
the sum of elements in column B whose field in column C is word:1 and
field in column A is word:7?'), we add the column name representations (described in Section \ref{selector}), $P$,  to $D$ to obtain column representations  $E$. This make the representation also sensitive to the column name.

In the second stage, we use $E$ to compute an attention over the
hidden states of the question RNN to get attention vector $G$ for each
column of the input table.
More concretely, we
compute the dot product between $E$ and the hidden states of the
question RNN to obtain scalar values. We then pass them through
softmax to obtain weighting factors for each hidden state. $G$ is the
weighted combination of the hidden states of the question RNN.

Finally, text match selection is done by:
\begin{align*}
 \mathit{text\_match}[m][k] = \mathit{sigmoid} \left(\sum \limits_{p=1}^d A[m][k][p] \cdot G[k][p] \right) \; \; \forall (m,k) \; \; m = 1, \ldots, M, k = 1, \ldots, K  
\end{align*}

Without loss of generality, let the first $K$ ($K \in [0, 1, \ldots,
  C]$) columns out of $C$ columns of the table contain text entries
while the remaining contain numeric entries. The row selector
variable now is given by:
\begin{align*}
 \mathit{row\_select_t[i]} = \alpha^{op}_t(\text{and}) & \mathit{and_t}[i]  + \alpha^{op}_t(\text{or}) \mathit{or_t}[i] + \alpha^{op}_t(\text{reset}) \mathit{reset_t}[i] 
 + \\ &\sum \limits_{j=K+1}^C \alpha^{col}_t(j) (\alpha^{op}_t(\text{greater}) g_t[i][j] + \alpha^{op}_t(\text{lesser}) l_t[i][j])
 + \\ &\sum \limits_{j=1}^K \alpha^{col}_t(j) (\alpha^{op}_t(\text{text\_match}) \mathit{text\_match}_t[i][j], \forall i = 1, \ldots, M
\end{align*}

The two-stage mechanism is required since in our experiments we find that simply averaging the vector representations fails to make the representation of the column specific enough to the question. Unless otherwise stated, our experiments are with input tables whose entries are only numeric and in that case the model does not contain the text match operation.
      
\subsection{History RNN}
\begin{figure}[h!]
  \includegraphics[scale=.28]{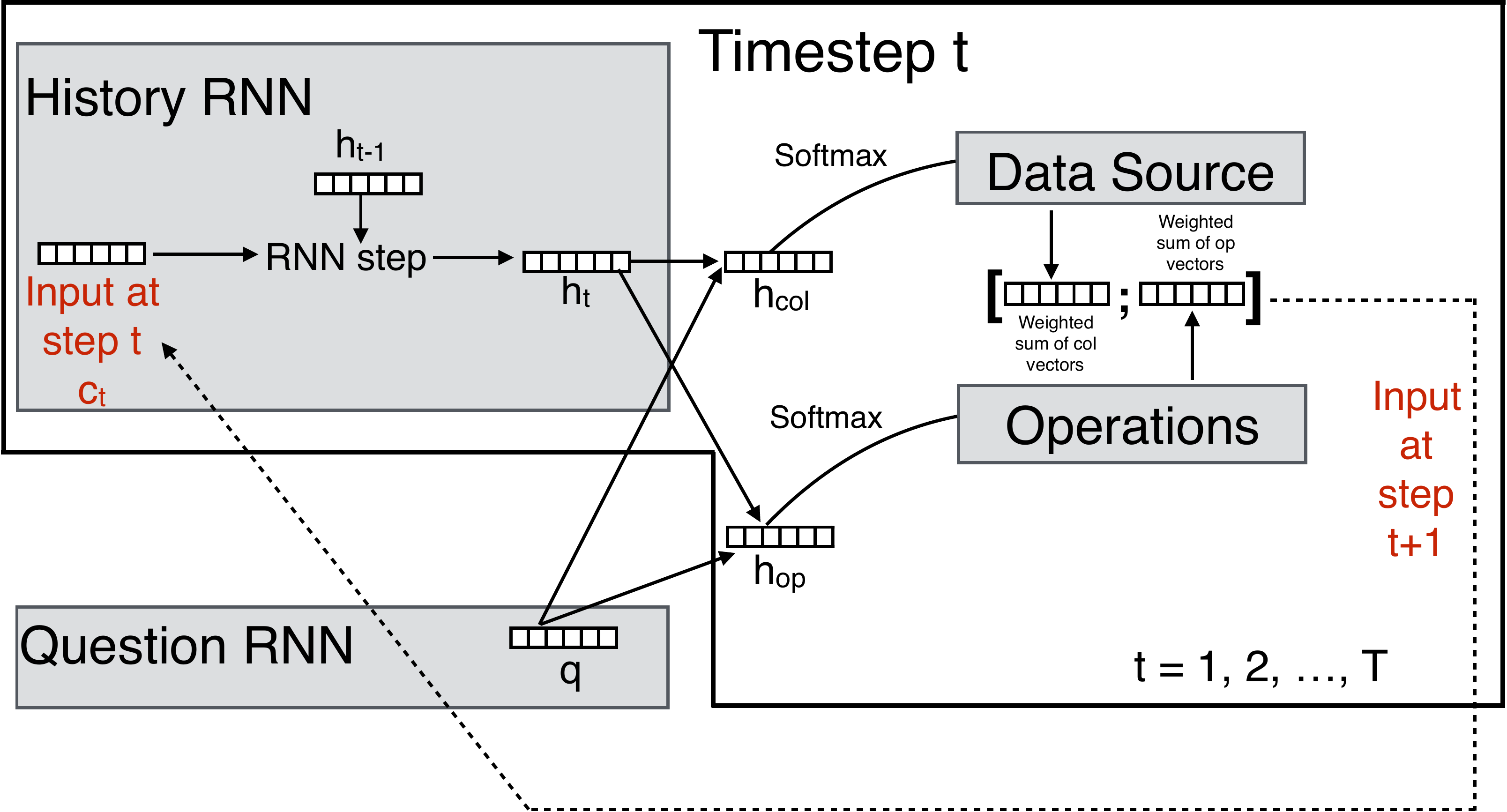}
  \caption{ The history RNN which
    helps in remembering the previous operations and data segments
    selected by the model. The dotted line indicates the input to the history RNN at step t+1.  } 
  \label{main:history}
\end{figure}

The history RNN keeps track of the previous operations and columns
selected by the selector module so that the model can induce compositional programs.  This information is encoded in the
hidden vector of the history RNN at time step $t$, $h_{t} \in
\mathbb{R}^{d}$. This helps the selector module to induce the
probability distributions over the operations and columns by taking into account the previous actions selected by the model. Figure
\ref{main:history} shows details of this component.
\label{history}

The input to the history RNN at time step $t$, $c_t \in
\mathbb{R}^{2d}$ is obtained by concatenating the weighted
representations of operations and column names with their
corresponding probability distribution produced by the selector at
step $t - 1$. More precisely:
\begin{align*}
c_t = [(\alpha^{op}_{t-1})^{T} U; (\alpha^{col}_{t-1})^{T} P]
\end{align*}
The hidden state of the history RNN at step $t$ is computed as:
\begin{align*}
h_{t} = \tanh(W^{\mathit{history}}[c_t; h_{t-1}]) , \forall i={1, 2, \ldots, Q}
\end{align*}
where $W^{\mathit{history}} \in \mathbb{R}^{d \times 3d}$ is the recurrent
matrix of the history RNN, 
and $h_t \in \mathbb{R}^{d}$ is the
current representation of the history. The history vector at time $t =
1$, $h_1$ is set to $[0]^{d}$.

\subsection{Training Objective}
\label{objective}
The parameters of the model include the parameters of the question
RNN, $W^{question}$, 
parameters of the history RNN,
$W^{history}$,
word embeddings $V(.)$, operation
embeddings $U$, operation selector and column selector matrices,
$W^{op}$ and $W^{col}$ respectively. During training, depending on
whether the answer is a scalar or a lookup from the table we
have two different loss functions.  

When the answer is a scalar, we use Huber loss \citep{huber} given by:
\begin{align*}
L_{scalar}(scalar\_answer_T, y)= 
\begin{cases}
    \frac{1}{2} a^2 , \text{if } a\leq \delta\\
    \delta a - \frac{1}{2} \delta^2,               \text{otherwise}
\end{cases}
\end{align*}
where $a = |\mathit{scalar\_answer}_{T} - y|$ is the absolute difference
between the predicted and true answer, and $\delta$ is the Huber
constant treated as a model hyper-parameter. In our experiments, we
find that using square loss makes training unstable while using the
absolute loss makes the optimization difficult near the
non-differentiable point.

When the answer is a list of items selected from the table, we convert
the answer to $y \in \{0,1\}^{M \times C}$, where $y[i, j]$ indicates
whether the element $(i, j)$ is part of the output. In this case we
use log-loss over the set of elements in the table given by:
\begin{align*}
 L_{\mathit{lookup}}(\mathit{lookup\_answer}_T, y) = -\frac{1}{MC} \sum \limits_{i=1}^{M} \sum \limits_{j=1}^{C} & \bigg(y[i, j]\log (\mathit{lookup\_answer}_T[i, j]) + \\ & (1 - y[i,j])\log (1 - \mathit{lookup\_answer}_T[i, j])\bigg)
\end{align*}

The training objective of the model is given by:
\begin{align*}
 L = \frac{1}{N}\sum \limits_{k=1}^{N} \bigg([n_k == True]L^{(k)}_{\mathit{scalar}} + [n_k == False]\lambda L^{(k)}_{\mathit{lookup}}\bigg)
\end{align*}
where $N$ is the number of training examples, $L^{(k)}_{\mathit{scalar}}$ and
$L^{(k)}_{\mathit{lookup}}$ are the scalar and lookup loss on $k^{th}$ example,
$n_k$ is a boolean random variable which is set to $\mathit{True}$ when the
$k^{th}$ example's answer is a scalar and set to $\mathit{False}$ when the
answer is a lookup, and $\lambda$ is a hyper-parameter of the model
that allows to weight the two loss functions appropriately.

At inference time, we replace the three softmax layers in the model
with the conventional maximum (hardmax) operation and the final output
of the model is either $scalar\_answer_T$ or $lookup\_answer_T$,
depending on whichever among them is updated after $T$ time steps. Algorithm \ref{algorithm}
gives a high-level view of Neural Programmer during inference.

\begin{algorithm}[h!]
\caption{High-level view of Neural Programmer during its inference
  stage for an input example.}
\label{algorithm}
\begin{algorithmic}[1]
\State {\bf Input}:  $\mathit{table} \in \mathbb{R}^{M \times C}$ and $\mathit{question}$
\State {\bf Initialize}:  $\mathit{scalar\_answer}_{0} = 0$, $\mathit{lookup\_answer}_{0} = {0}^{M \times C}$, $\mathit{row\_select}_0 = {1} ^ {M}$, history vector at time $t=0$, $h_0 = 0^{d}$ and input to history RNN at time $t=0$, $c_0 = 0^{2d}$
\State{\bf Preprocessing}: Remove numbers from $\mathit{question}$ and store them in a list along with the words that appear to the left of it. The tokens in the input question are $\{w_{1}, w_{2}, \ldots, w_{Q}\}$. 
\State {\bf Question Module}:  Run question RNN on the preprocessed question to get question representation $q$ and list of hidden states $z_1, z_2, \ldots, z_Q$
\State {\bf Pivot numbers}: $\mathit{pivot_g}$ and $\mathit{pivot_l}$ are computed using  hidden states from question RNN and operation representations $U$
\For{$t=1,2, \ldots ,T$ }
\State Compute history vector $h_t$ by passing input $c_t$ to the history RNN
\State{\bf  Operation selection} using $q$, $h_t$ and operation representations $U$
\State{\bf  Data selection} on $table$ using $q$, $h_t$ and column representations $V$
\State Update $\mathit{scalar\_answer_t}$, $\mathit{lookup\_answer_t}$ and $\mathit{row\_select}_t$ using the selected operation and column
\State Compute  input to the history RNN at time $t+1$, $c_{t+1}$
\EndFor
\State {\bf Output}: $\mathit{scalar\_answer}_{T}$ or $\mathit{lookup\_answer}_{T}$ depending on whichever of the two is updated at step $T$
\end{algorithmic}
\end{algorithm}

\section{Experiments}
Neural Programmer is faced with many challenges, specifically: 1) can
the model learn the parameters of the different modules with delayed
supervision after $T$ steps? 2) can it exhibit compositionality by
generalizing to unseen questions? and 3) can the question module
handle the variability and ambiguity of natural language? In our
experiments, we mainly focus on answering the first two questions
using synthetic data. Our reason for using synthetic data is that it
is easier to understand a new model with a synthetic dataset. We can
generate the data in a large quantity, whereas the biggest real-word
semantic parsing datasets we know of contains only about 14k training
examples \citep{PasupatL15} which is very small by neural network
standards. In one of our experiments, we introduce simple word-level
variability to simulate one aspect of the difficulties in dealing with
natural language input.

\subsection{Data}
We generate question, table and answer triples using a synthetic
grammar. Tables \ref{type-questions} and \ref{3-questions} (see
Appendix) shows examples of question templates from the synthetic
grammar for single and multiple columns respectively. The elements in
the table are uniformly randomly sampled from [-100, 100] and [-200,
  200] during training and test time respectively. The number of rows
is sampled randomly from [30, 100] in training while during prediction
the number of rows is 120. Each question in the test set is unique,
i.e., it is generated from a distinct template. We use the following
settings:

{\bf Single Column:} We first perform experiments with a single column
that enables $23$ different question templates which can be
answered using $4$ time steps.

{\bf Many Columns:} We increase the difficulty by experimenting with
multiple columns (max\_columns = 3, 5 or 10). During training, the
number of columns is randomly sampled from (1, max\_columns) and at
test time every question had the maximum number of columns used during
training.

{\bf Variability:} {To simulate  one aspect of  the difficulties in dealing with natural language input, we consider multiple ways to refer to the same operation (Tables \ref{wv} and \ref{wv:question}).}

{\bf Text Match:} Now we consider cases where some columns in the
input table contain text entries. We use a small vocabulary of 10
words and fill the column by uniformly randomly sampling from them. In
our first experiment with text entries, the table always contains two
columns, one with text and other with numeric entries (Table
\ref{text-match}). In the next experiment, each example can have up to
3 columns containing numeric entries and up to 2 columns containing
text entries during training. At test time, all the examples contain 3
columns with numeric entries and 2 columns with text entries.

\subsection{Models}
In the following, we benchmark the performance of Neural Programmer on
various versions of the table-comprehension dataset. We slowly
increase the difficulty of the task by changing the table properties
(more columns, mixed numeric and text entries) and question properties
(word variability). After that we discuss a comparison between Neural
Programmer, LSTM, and LSTM with Attention.
\subsubsection{Neural Programmer}
We use $4$ time steps in our experiments ($T = 4$). Neural Programmer
is trained with mini-batch stochastic gradient descent with Adam
optimizer~\citep{KingmaB14}. The parameters are initialized uniformly
randomly within the range [-0.1, 0.1]. In all experiments, we set the
mini-batch size to $50$, dimensionality $d$ to $256$, the initial
learning rate and the momentum hyper-parameters of Adam to their
default values~\citep{KingmaB14}. We found that it is extremely useful
to add random Gaussian noise to our gradients at every training
step. This acts as a regularizer to the model and allows it to
actively explore more programs. We use a schedule inspired from
\citet{WellingT11}, where at every step we sample a Gaussian of $0$
mean and variance$=\text{curr\_step}^{-0.55}$.

To prevent exploding gradients, we perform gradient clipping by
scaling the gradient when the norm exceeds a threshold
\citep{Graves13}. The threshold value is picked from [1, 5, 50].  We tune
the $\epsilon$ hyper-parameter in Adam from [1e-6, 1e-8], the Huber
constant $\delta$ from [10, 25, 50] and $\lambda$ (weight between two
losses) from [25, 50, 75, 100] using grid search. While performing
experiments with multiple random restarts we find that the performance
of the model is stable with respect to $\epsilon$ and gradient
clipping threshold but we have to tune $\delta$ and $\lambda$ for the
different random seeds.

\begin{center}
    \begin{table}[h!]
    \begin{tabular}{| l | l | l | l |}
    \hline
    Type & No. of Test Question Templates   & Accuracy  & \% seen test \\ \hline \hline
    Single Column & 23  & 100.0 & 100 \\ \hline \hline
    3 Columns & 307  & 99.02 & 100 \\ \hline
    5 Columns & 1231 & 99.11 & 98.62 \\ \hline
    10 Columns & 7900 & 99.13 & 62.44 \\ \hline \hline
    Word Variability on 1 Column  & 1368 & 96.49 & 100 \\ \hline
    Word Variability on 5 Columns  & 24000 & 88.99 & 31.31 \\ \hline \hline 
    Text Match on 2 Columns  & 1125 & 99.11 & 97.42 \\ \hline
    Text Match on 5 Columns  & 14600 & 98.03 & 31.02 \\ \hline 
    \end{tabular}
    \caption{Summary of the performance of Neural Programmer on
      various versions of the synthetic table-comprehension task. The prediction of the model
      is considered correct if it is equal to the correct answer up to
      the first decimal place. The last column indicates the
      percentage of question templates in the test set that are observed during
      training.  The unseen question templates generate questions containing sequences of
      words that the model has never seen before. The model can generalize to unseen question templates which is
      evident in the 10-columns, word variability on  5-columns and text match on 5 columns experiments.  This indicates that
      Neural Programmer is a powerful compositional model since solving unseen question templates requires performing a sequence of actions that it has never done during training.}
    \label{results}
    \end{table}
\end{center}

The training set consists of $50,000$ triples in all our
experiments. Table \ref{results} shows the performance of Neural
Programmer on synthetic data experiments. In single column
experiments, the model answers all questions correctly which we
manually verify by inspecting the programs induced by the model. In
many columns experiments with 5 columns, we use a bidirectional RNN
and for 10 columns we additionally perform attention
\citep{BahdanauCB14} on the question at every time step using the
history vector. The model is able to generalize to unseen question
templates which are a considerable fraction in our ten columns
experiment. This can also be seen in the word variability experiment
with 5 columns and text match experiment with 5 columns where more
than two-thirds of the test set contains question templates that are
unseen during training. This indicates that Neural Programmer is a
powerful compositional model since solving unseen question templates
requires inducing programs that do not appear during training. Almost
all the errors made by the model were on questions that require the
\emph{difference operation} to be used.  Table \ref{parse} shows examples of
how the model selects the operation and column at every time step for
three test questions.

\begin{center}
    \begin{table}[h!]
     \small
    \begin{tabular}{| c | c | c | c | c | c | c | c |}
    \hline
    {\multirow{2}{*} {Question}} & {\multirow{2}{*} {t}} & Selected & Selected & $pivot_g$ & $pivot_l$  & Row  \\
                                 &                          & Op & Column &  &  & select                                    \\ \hline \hline
    greater 50.32 C and lesser 20.21 E sum H  & 1 & Greater    & C &  {\multirow{4}{*} {50.32}} &  {\multirow{4}{*} {20.21}}  & $g_1$ \\
    {\it What is the sum of numbers in column H }	& 2  & Lesser    & E &  &   & $l_2$ \\
    {\it whose  field in column C is greater than 50.32} & 3 & And   & - &  &  & $and_3$ \\
    {\it and  field in Column E is lesser than 20.21.}	& 4 & Sum    & H &  &   & $[0]^M$ \\ \hline \hline
    lesser -80.97 D or greater 12.57 B print F  & 1 & Lesser    & D & {\multirow{4}{*} {12.57}} & {\multirow{4}{*} {-80.97}}  & $l_1$ \\
    {\it Print elements in column F}	& 2  & Greater    & B &  &   & $g_2$ \\
    {\it whose  field in column D  is lesser than -80.97} & 3 & Or    & - & &  & $or_3$ \\
    {\it or  field in Column B is greater than 12.57.}	& 4 & Assign    & F &  &   & $[0]^M$ \\ \hline \hline
    sum A diff count & 1 & Sum    & A & {\multirow{4}{*} {-1}} & {\multirow{4}{*} {-1}}  & $[0]^M$ \\
    {\it What is the difference} & 2  & Reset  & - & &  & $[1]^M$ \\
    {\it between sum of elements in} & 3 & Count  & - & &  & $[0]^M$ \\
    {\it column A and number of rows} & 4 & Diff  & - & & & $[0]^M$ \\ \hline							
    \end{tabular}
    \caption{Example outputs from the model for $T=4$ time steps on
      three questions in the test set. We show the synthetically
      generated question along with its natural language
      translation. For each question, the model takes 4 steps and at
      each step selects an operation and a column. The pivot numbers
      for the comparison operations are computed before performing the
      4 steps. We show the selected columns in cases during which the
      selected operation acts on a particular column.}
    \label{parse}
    \end{table}
\end{center}

\begin{figure}[h!]
 \centering
  \includegraphics[scale=.35]{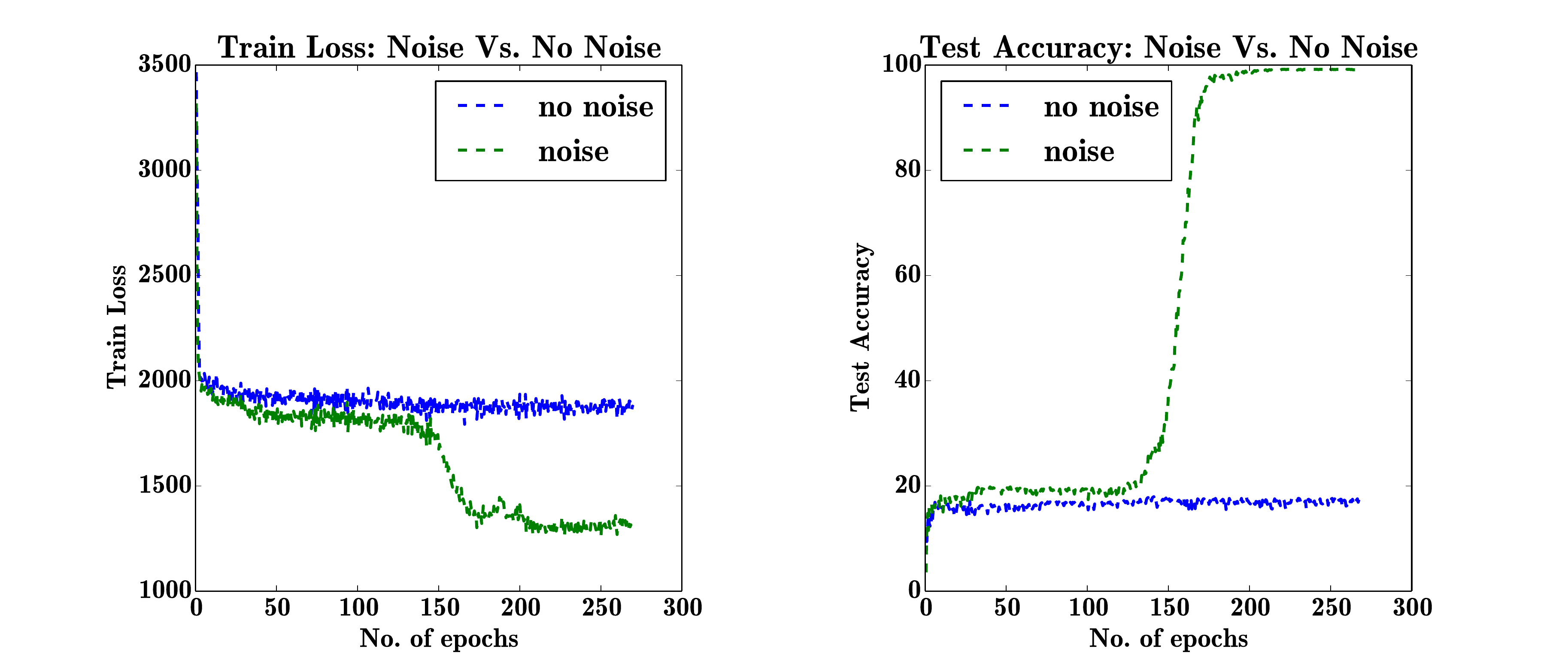}
 \caption{The effect of adding random noise to the gradients versus
   not adding it in our experiment with 5 columns when all    hyper-parameters are the same. The models trained with noise    generalizes almost always better.}
  \label{noise}
\end{figure}

Figure \ref{noise} shows an example of the effect of adding random noise to the gradients in our experiment with $5$ columns.

\subsubsection{Comparison to LSTM and LSTM with Attention}
We apply a three-layer sequence-to-sequence LSTM recurrent network
model \citep{Hochreiter:1997,SutskeverVL14} and LSTM model with
attention \citep{BahdanauCB14}. We explore multiple attention heads
(1, 5, 10) and try two cases, placing the input table before and after
the question. We consider a \emph{simpler version} of the single
column dataset with only questions that have scalar answers. The
number of elements in the column is uniformly randomly sampled from
$[4, 7]$ while the elements are sampled from $[-10, 10]$.  The best
accuracy using these models is close to $80\%$ in spite of relatively
easier questions and supplying fresh training examples at every
step. When the scale of the input numbers is changed to $[-50, 50]$ at
test time, the accuracy drops to $30\%$.

Neural Programmer solves this task and achieves $100\%$ accuracy using
$50,000$ training examples. Since hardmax operation is used at test
time, the answers (or the program induced) from Neural Programmer is
invariant to the scale of numbers and the length of the input.

\section{Related Work}
Program induction has been studied in the context of semantic parsing
\citep{zelle:aaai96,Zettlemoyer05learningto,Liang:2011} in natural
language processing. \citet{PasupatL15} develop a semantic parser with
a hand engineered grammar for question answering on tables with
natural language questions. Methods such as
\citet{piantadosi2008bayesian,Eisenstein:2009,clarke-EtAl:2010:CONLL}
learn a compositional semantic model without hand engineered
compositional grammar, but still requiring a hand labeled lexical
mapping of words to the operations. \citet{Poon13} develop an
unsupervised method for semantic parsing, which requires many
pre-processing steps including dependency parsing and mapping from
words to operations. \citet{liang2010learning} propose an hierarchical
Bayesian approach to learn simple programs.

There has been some early work in using neural networks for learning
context free grammar
\citep{Das92learningcontext-free,Das92usingprior,zeng94discrete} and
context sensitive grammar
\citep{Steijvers96arecurrent,Gers01lstmrecurrent} for small
problems. \citet{NeelakantanRM15,LinLLSRL15} learn simple Horn clauses
in a large knowledge base using RNNs.  Neural networks have also been
used for Q\&A on datasets that do not require complicated arithmetic
and logic reasoning \citep{BordesCW14,IyyerBCSD14,sukhbaatar2015weakly,DBLP:journals/corr/PengLLW15,DBLP:journals/corr/HermannKGEKSB15}. While there has
been lot of work in augmenting neural networks with additional memory
\citep{Das92learningcontext-free,Schmidhuber:93,Hochreiter:1997,GravesWD14,WestonCB14,KumarISBEPOGS15,JoulinM15},
 we are not aware of any other work that
augments a neural network with a set of operations to enhance
complex reasoning capabilities.

After our work was submitted to ArXiv,  Neural Programmer-Interpreters \citep{reed2015}, a method that learns to induce programs with supervision of the entire program was proposed. This was followed by  Neural Enquirer \citep{YinLLK15}, which similar to our work tackles the problem of synthetic table QA. However, their method achieves perfect accuracy only when given supervision of the entire program. Later, dynamic neural module network \citep{jacob} was proposed for question answering which uses syntactic supervision in the form of dependency trees. 

\section{Conclusions}
We develop Neural Programmer, a neural network model augmented with a
small set of arithmetic and logic operations to perform complex arithmetic and logic 
reasoning. The model can be trained in
an end-to-end fashion using backpropagation to induce programs requiring much lesser sophisticated
human supervision than prior work. It is a general model for program induction broadly applicable
across different domains, data sources and languages. Our experiments indicate
that the model is capable of learning with delayed supervision and
exhibits powerful compositionality.


\paragraph{Acknowledgements}
We sincerely thank Greg Corrado, Andrew Dai, Jeff Dean, Shixiang Gu, Andrew
McCallum, and Luke Vilnis for their suggestions and the Google Brain
team for the support.

\bibliography{neural_programmer}
\bibliographystyle{iclr2016_conference}
\newpage
\section*{Appendix}
\begin{center}
    \begin{table}[h]
    \begin{tabular}{ | l |}
    \hline 
      sum \\
      count \\
      print \\
      greater [number] sum \\
      lesser [number] sum  \\
      greater [number] count  \\
      lesser [number] count \\
      greater [number] print \\
      lesser [number] print  \\
      greater [number1] and lesser [number2] sum \\
      lesser [number1] and greater [number2] sum \\
      greater [number1] or lesser [number2] sum \\
      lesser [number1] or greater [number2] sum \\
      greater [number1] and lesser [number2] count \\
      lesser [number1] and greater [number2] count \\
      greater [number1] or lesser [number2] count \\
      lesser [number1] or greater [number2] count \\
      greater [number1] and lesser [number2] print \\
      lesser [number1] and greater [number2] print \\
      greater [number1] or lesser [number2] print \\
      lesser [number1] or greater [number2] print \\
      sum diff count \\
      count diff sum \\
      \hline
    \end{tabular}
    \caption{23  question templates for single column experiment. We have four categories of questions: 1) simple aggregation (sum, count) 2) comparison (greater, lesser) 3) logic (and, or) and, 4) arithmetic (diff). We first sample the categories uniformly randomly and each program within a category is equally likely. In the word variability experiment with $5$ columns we sampled from the set of all programs uniformly randomly since greater than $90\%$ of the test questions were unseen during training  using the other procedure.}
    \label{type-questions}
    \end{table}
\end{center}

\begin{center}
    \begin{table}[h]
    \begin{tabular}{ | l |}
    \hline
      greater [number1] A and lesser [number2] A  sum A\\
      greater [number1] B and lesser [number2] B  sum B\\
      greater [number1] A and lesser [number2] A  sum B\\
      greater [number1] A and lesser [number2] B  sum A\\
      greater [number1] B and lesser [number2] A  sum A\\
      greater [number1] A and lesser [number2] B  sum B\\
      greater [number1] B and lesser [number2] B  sum A\\
      greater [number1] B and lesser [number2] B  sum A\\
      \hline
    \end{tabular}
    \caption{8  question templates of type ``greater [number1] and lesser [number2] sum'' when there are 2 columns.}
    \label{3-questions}
    \end{table}
\end{center}

\begin{center}
    \begin{table}[h]
    \begin{tabular}{ | l | l |}
    \hline
    sum & sum, total, total of, sum of \\
    count & count, count of, how many \\
    greater & greater,  greater than, bigger, bigger than, larger, larger than \\
    lesser & lesser, lesser than, smaller, smaller than, under \\
    assign & print, display, show \\
    difference  & difference, difference between \\  
    \hline
    \end{tabular}
    \caption{Word variability, multiple ways to refer to the same operation.}
    \label{wv}
    \end{table}
\end{center}

\begin{center}
    \begin{table}[h]
    \begin{tabular}{ | l | l |}
    \hline
    greater [number] sum \\
    greater [number] total \\
    greater [number] total of \\ 
    greater [number] sum of \\
    
    greater than [number] sum \\
    greater than [number] total \\
    greater than [number] total of \\ 
    greater than [number] sum of \\
    
    bigger [number] sum \\
    bigger [number] total \\
    bigger [number] total of \\ 
    bigger [number] sum of \\
    
    bigger than [number] sum \\
    bigger than [number] total \\
    bigger than [number] total of \\ 
    bigger than [number] sum of \\
    
    larger [number] sum \\
    larger [number] total \\
    larger [number] total of \\ 
    larger [number] sum of \\
    
    larger than [number] sum \\
    larger than [number] total \\
    larger than [number] total of \\ 
    larger than [number] sum of \\
    \hline
    \end{tabular}
    \caption{24 questions templates for questions of type ``greater [number] sum'' in the single column word variability experiment.}
    \label{wv:question}
    \end{table}
\end{center}

\begin{center}
    \begin{table}[h]
    \begin{tabular}{ | l |}
    \hline
    word:0 A sum B \\
    word:1 A sum B \\
    word:2 A sum B \\
    word:3 A sum B \\
    word:4 A sum B \\
    word:5 A sum B \\
    word:6 A sum B \\
    word:7 A sum B \\
    word:8 A sum B \\
    word:9 A sum B \\
    \hline
    \end{tabular}
    \caption{10 questions templates for questions of type ``[word] A sum B'' in the two columns text match experiment.}
    \label{text-match}
    \end{table}
\end{center}

\end{document}